%% file: main.tex
\documentclass[10pt,twocolumn,letterpaper]{article}

\usepackage[accsupp]{axessibility}
\usepackage[pagenumbers]{wacv} 

\input{preamble}

\definecolor{wacvblue}{rgb}{0.21,0.49,0.74}
\usepackage[pagebackref,breaklinks,colorlinks,allcolors=wacvblue]{hyperref}
\usepackage{graphicx,verbatim}
\usepackage{bm}
\usepackage{booktabs}
\usepackage{amsmath,amssymb,amsfonts}
\usepackage{algorithm}
\usepackage{algorithmic}
\usepackage{multirow}
\usepackage[normalem]{ulem}
\usepackage{textcomp}
\usepackage{xcolor}
\usepackage{colortbl}
\usepackage{caption}
\usepackage{subcaption}
\def\wacvPaperID{1208} 
\def\confName{WACV}
\def\confYear{2027}

\title{Findings-aware Language-Image Pretraining for CT Volume Interpretation}

\author{Takuro Shimaya, Junya Inoh, Masahiro Saiko\\
Biometrics Research Labs, NEC corporation, Kawasaki, Japan\\
{\tt\small takuro-shimaya@nec.com}
}

\begin{document}
\maketitle
\input{sec/0_abstract}    
\input{sec/1_intro}
\input{sec/2_method}
\input{sec/3_results}
\input{sec/4_conclusions}
{
    \small
    \bibliographystyle{ieeenat_fullname}
    \bibliography{main}
}

\end{document}


\onecolumn

\title{Supplementary text for ``Findings-aware Language-Image Pretraining for CT Volume Interpretation''}  
\author{}

\maketitle
\thispagestyle{empty}
\appendix


\section*{Supplementary tables}
\begin{table}[h]
\centering
\caption{P-values for the comparison of different models on various metrics BEFORE Holm correction.}
\resizebox{\textwidth}{!}{%
\begin{tabular}{l||cc|ccccc|cc}
     & \multicolumn{2}{c}{Report gen. metrics} & \multicolumn{5}{c}{GREEN subcategory} & \multicolumn{2}{c}{Zero-shot AUC} \\
    models & BLEU-4 & GREEN & g-1 & g-2 & g-3 & g-4 & g-5 & CT-RATE & RadChestCT\\
    \hline\hline
    FiLIP vs ImageNet & $8.5\times 10^{-4}$ & $7.9\times 10^{-4}$ & $<10^{-4}$ & $1.1\times 10^{-1}$ & $<10^{-4}$ & $4.1\times 10^{-1}$ & $4.5\times 10^{-2}$ & - & -  \\
    \hline
    FiLIP vs CT-CLIP & - & - & - & - & - & - & - & $<10^{-4}$ & $<10^{-4}$  \\
    \hline
    FiLIP vs fVLM & $<10^{-4}$ & $1.9\times 10^{-3}$ & $<10^{-4}$ & $3.0\times 10^{-1}$ & $1.3\times 10^{-3}$ & $1.5\times 10^{-1}$ & $7.6\times 10^{-1}$ & $<10^{-4}$ & $1.1\times 10^{-3}$ \\
    \hline
    FiLIP vs SimCroP & $1.8\times 10^{-1}$ & $6.5\times 10^{-2}$ & $<10^{-4}$ & $2.6\times 10^{-1}$ & $<10^{-4}$ & $8.9\times 10^{-3}$ & $1.7\times 10^{-1}$ & $<10^{-4}$ & $<10^{-4}$ \\
    \hline
    \modelcell{FiLIP (w/ BioLORD) vs}{FiLIP (w/ CXR-BERT)} & - & - & - & - & - & - & - & $<10^{-4}$ & - \\
    \hline
    \modelcell{FiLIP (w/ BioLORD) vs}{FiLIP (w/ text-embedding-3-large)} & - & - & - & - & - & - & - & $<10^{-4}$ & - \\
    \hline
    \modelcell{FiLIP (w/ GPT-4.1) vs}{FiLIP (w/ gpt-oss-120b)} & - & - & - & - & - & - & - & $4.4\times 10^{-1}$ & - \\
    \hline
\end{tabular}
}
\end{table}

\begin{table}[h]
\centering
\caption{P-values for the comparison of different models on various metrics AFTER Holm correction.
The \textbf{\underline{bold and underlined values}} indicate statistical significance at a 5\% significance level.}
\resizebox{\textwidth}{!}{%
\begin{tabular}{l||cc|ccccc|cc}
     & \multicolumn{2}{c}{Report gen. metrics} & \multicolumn{5}{c}{GREEN subcategory} & \multicolumn{2}{c}{Zero-shot AUC} \\
    models & BLEU-4 & GREEN & g-1 & g-2 & g-3 & g-4 & g-5 & CT-RATE & RadChestCT\\
    \hline\hline
    FiLIP vs ImageNet & $\sig{1.4\times 10^{-2}}$ & $\sig{1.4\times 10^{-2}}$ & $\sig{<10^{-4}}$ & $>0.99$ & $\sig{6.5\times 10^{-4}}$ & $>0.99$ & $4.9\times 10^{-1}$ & - & -  \\
    \hline
    FiLIP vs CT-CLIP & - & - & - & - & - & - & - & $\sig{<10^{-4}}$ & $\sig{<10^{-4}}$  \\
    \hline
    FiLIP vs fVLM & $\sig{<10^{-4}}$ & $\sig{2.4\times 10^{-2}}$ & $\sig{<10^{-4}}$ & $>0.99$ & $\sig{1.8\times 10^{-2}}$ & $>0.99$ & $7.6\times 10^{-1}$ & $\sig{<10^{-4}}$ & $\sig{1.7\times 10^{-2}}$ \\
    \hline
    FiLIP vs SimCroP & $>0.99$ & $6.5\times 10^{-1}$ & $\sig{1.7\times 10^{-4}}$ & $>0.99$ & $\sig{<10^{-4}}$ & $1.1\times 10^{-1}$ & $>0.99$ & $\sig{<10^{-4}}$ & $\sig{<10^{-4}}$ \\
    \hline
    \modelcell{FiLIP (w/ BioLORD) vs}{FiLIP (w/ CXR-BERT)} & - & - & - & - & - & - & - & $\sig{<10^{-4}}$ & - \\
    \hline
    \modelcell{FiLIP (w/ BioLORD) vs}{FiLIP (w/ text-embedding-3-large)} & - & - & - & - & - & - & - & $\sig{1.4\times 10^{-2}}$ & - \\
    \hline
    \modelcell{FiLIP (w/ GPT-4.1) vs}{FiLIP (w/ gpt-oss-120B)} & - & - & - & - & - & - & - & $8.9\times 10^{-1}$ & - \\
    \hline
\end{tabular}
}
\end{table}

\clearpage
\begin{table}[h]
\centering
\caption{Correspondence between finding labels and organs, which is equivalent to that adopted in fVLM.}
\begin{tabular}{ll}
Finding label & Organ \\
\hline\hline
Arterial wall calcification & aorta \\
\hline
\multirow[t]{3}{*}{Cardiomegaly} & \multirow[t]{3}{*}{heart} \\
\multicolumn{1}{l}{Pericardial effusion} & \\
\multicolumn{1}{l}{Coronary artery wall calcification} & \\
\hline
Hiatal hernia & esophagus \\
\hline
\multirow[t]{11}{*}{Emphysema} & \multirow[t]{11}{*}{lung} \\
\multicolumn{1}{l}{Atelectasis} & \\
\multicolumn{1}{l}{Lung nodule} & \\
\multicolumn{1}{l}{Lung opacity} & \\
\multicolumn{1}{l}{Pulmonary fibrotic sequela} & \\
\multicolumn{1}{l}{Pleural effusion} & \\
\multicolumn{1}{l}{Mosaic attenuation pattern} & \\
\multicolumn{1}{l}{Peribronchial thickening} & \\
\multicolumn{1}{l}{Consolidation} & \\
\multicolumn{1}{l}{Bronchiectasis} & \\
\multicolumn{1}{l}{Interlobular septal thickening} & \\
\hline
\end{tabular}
\end{table}

\section*{Supplementary note for the calculation of the inter-image triplet loss}
The average and std of unique phrases present only in the anchor, i.e., the number of element in $U^o_{i,j}$ (Eq.~(5)) is $4.6\pm0.4$~/anchor when $S_\mathrm{thr}=0.6$.
We set the batch size to 56, and the number of unique phrases in a batch is $4.6\times 55\approx 254$ on average.
This is the effective number of negative samples for the inter-image triplet loss.

Regarding the time required for training, the average and std time per epoch was $1,743\pm 22$ sec on NVIDIA A100 (80GB) * 8 GPUs, while $440\pm 39$ sec without the inter image triplet loss and $597\pm 34$ sec for fVLM without co-teaching.
The inter-image triplet loss requires $O(|B|^2)$ computational cost as shown in Eq.~(8), and is thus more computationally expensive than the intra-image focal similarity loss.
However, considering that fVLM with co-teaching (whose implementation is not publicly available) requires training two models simultaneously, including the text encoder, the computational cost for training FiLIP is not substantially different from that of fVLM.

\section*{Supplementary note for the architecture for radiology report generation tasks}
We trained an end-to-end model for radiology report generation using CT-RATE.
The pretrained image encoder was connected to the BLIP text decoder through cross-attention mediated by a visual projector, and the entire network was fine-tuned (Fig.~4b).
The visual projector sequentially applies a linear layer, a GELU activation, a second linear layer, positional encoding, layer normalization, and 10\% dropout, followed by 3D adaptive average pooling, while preserving the embedding dimension of the image features.
We then obtained a feature map with 768 channels and spatial dimensions of $4\times4\times8$ along the right-to-left, anterior-to-posterior, and craniocaudal axes, respectively.
During training, we used nearly the same hyperparameter settings as for image-encoder pretraining, except that the learning rate was reduced from 1e-4 to 1e-5.
For the loss function, we used the same formulation as that used for the text decoder in BLIP.


%% file: preamble.tex
\newcommand{\figref}[1]{Fig.~\ref{#1}}

\newcommand{\tabref}[1]{Table~\ref{#1}}
\newcommand{\equaref}[1]{Eq.~\eqref{#1}}
\newcommand{\lrparen}[1]{\left( #1 \right)}

\newcommand{\lrbrace}[1]{\left\{ #1 \right\}}

%% file: sec/0_abstract.tex
\begin{abstract}
Computed tomography (CT) provides diverse patient information and various clinical tasks, for which numerous deep learning models have been investigated.
Recent years have seen the development of vision-language foundation models that can comprehensively extract clinical features, driven by representation learning that combines CT volumes with radiology reports.
Existing approaches typically adopt Contrastive Language-Image Pre-training (CLIP)-based representation learning, where pairs from different cases are, in principle, treated as negative pairs such that the volume and report embeddings are pushed apart.
In the biomedical imaging, however, treating different cases as negative pairs is not always appropriate, particularly when clinical similarity exists.
In this study, we propose a supervised representation learning framework grounded in clinical knowledge from language foundation models, rather than the weak supervision used in CLIP.
Specifically, we utilized a large language model to extract abnormal finding phrases from reports.
We then explicitly constructed the target features that the image encoder should output, as the semantic representation of the extracted findings using a biomedical text embedding model.
Our Findings-aware Language-Image Pre-training (FiLIP) outperformed weakly supervised approaches in multi-label zero-shot finding classification in both in-domain and external-domain evaluations, and also proved effective when fine-tuned for the downstream task of radiology report generation.
\end{abstract}

%% file: sec/1_intro.tex
\section{Introduction}
\label{sec:intro}

Computed tomography (CT) is one of the most widely used medical imaging modalities, providing diverse clinical information across multiple organs.
CT imaging contributes to a broad range of clinical tasks, \eg, by supplying critical information for disease diagnosis, treatment planning, and digital twin construction \cite{Seeram2022,Zhao2025}.
Accordingly, numerous deep learning models have been developed, and translation to clinical practice has become increasingly common, particularly in computer-aided diagnosis \cite{Koetzier2023,Kleber2024,Gao2025}.
\begin{figure}[t]
    \centerline{\includegraphics[width=0.86\linewidth]{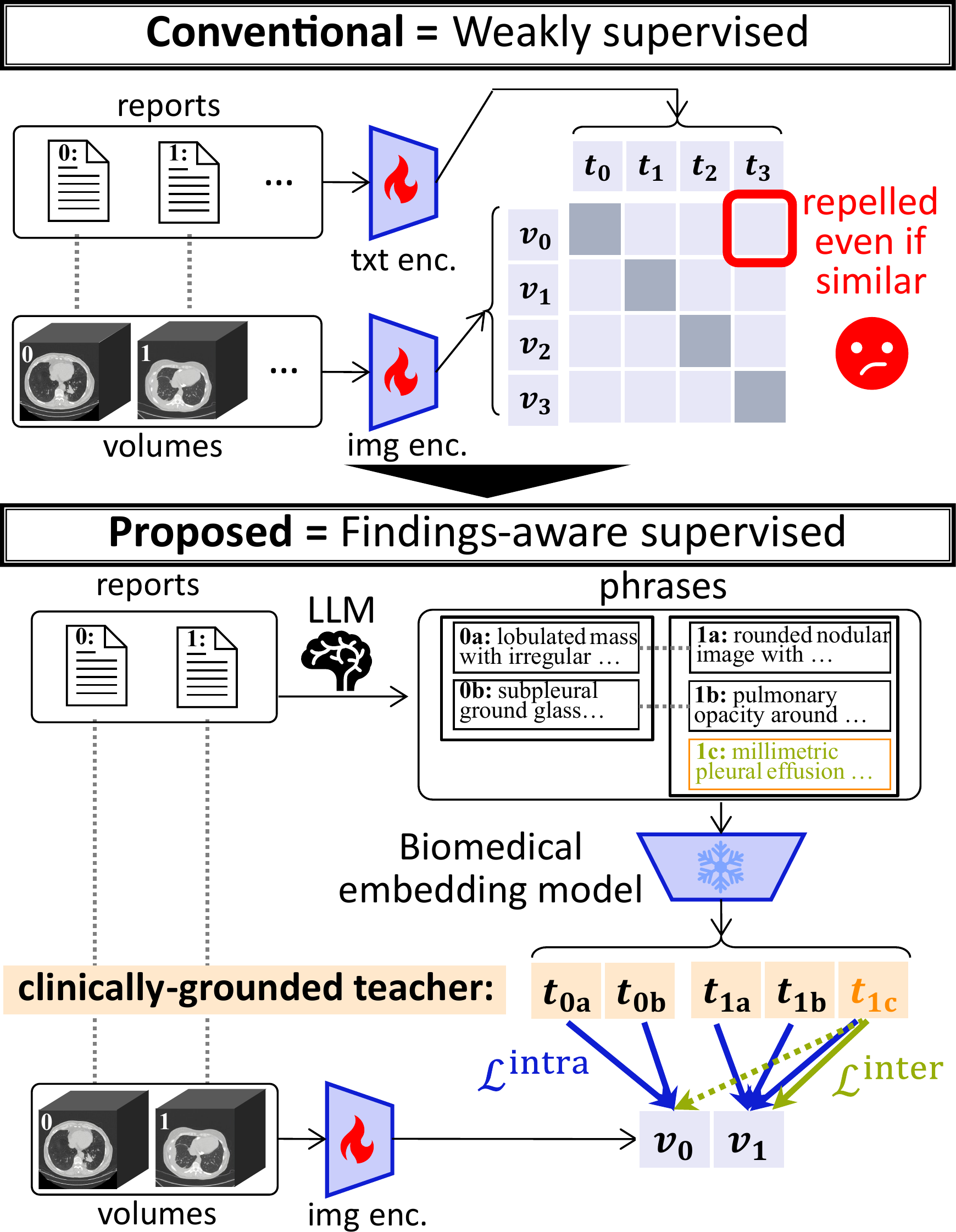}}
    \caption{Schematic comparison of conventional methods and the proposed FiLIP.
    While contrastive learning may treat similar cases as negative pairs, our method directly supervises the model using text embeddings of extracted finding phrases from reports.
    We leverage a pretrained biomedical embedding model and an LLM as external clinical knowledge sources.
    }
    \label{fig:1-1}
\end{figure}

Conventional deep learning has relied on time-consuming manual annotations provided by medical experts.
To reduce this burden, there has been growing interest in developing foundation models using large-scale medical imaging datasets.
Unlike task-specific models, foundation models learn broad clinical representations and can support applications such as zero-shot diagnosis without manual annotation \cite{Paschali2025, Lin2025}.
Accordingly, large-scale pretraining has been explored with datasets across multiple modalities such as RadImageNet \cite{Mei2022} and ROCOv2 \cite{Ruckert2024}.
In parallel, modality-specific vision-language foundation models have been developed for chest X-ray \cite{Chen2024} and histopathology \cite{Bilal2025}.
This progress has been largely driven by Contrastive Language-Image Pretraining (CLIP) \cite{Radford2021,Zhang2024}, which has achieved substantial success in computer vision.

This trend also applies to CT.
Three-dimensional volumes contain various abnormalities within a small fraction of slices, making manual annotation extremely burdensome.
Recently, the release of CT-RATE \cite{Hamamci2026}, an open dataset that pairs CT volumes with corresponding radiology reports, has accelerated the development of vision-language foundation models for CT \cite{Ryu2025,Hamamci2026,Cao2024,Blankemeier2026,Shui2025,Wang2025,Wald2025}.
Conventional foundation models designed for two-dimensional slices have limited applicability because clinical practice often requires three-dimensional interpretation.
To address this problem, CT-CLIP \cite{Hamamci2026}, which applies CLIP directly to CT volumes, achieved zero-shot classification of various findings and case retrieval.
Further improvement has been achieved by fVLM \cite{Shui2025}, which performs contrastive learning on an organ-wise basis; SimCroP \cite{Wang2025}, which learns representations at the sentence level in reports; and ViSD-Boost \cite{Cao2025} and CORIPLI \cite{Wald2025}, which combine vision-language alignment with additional loss functions such as autoencoding losses.
While the above methods involve minor modifications, they are still trained primarily using CLIP-style vision-language contrastive learning.
\begin{figure*}[t]
    \centerline{\includegraphics[width=0.9\linewidth]{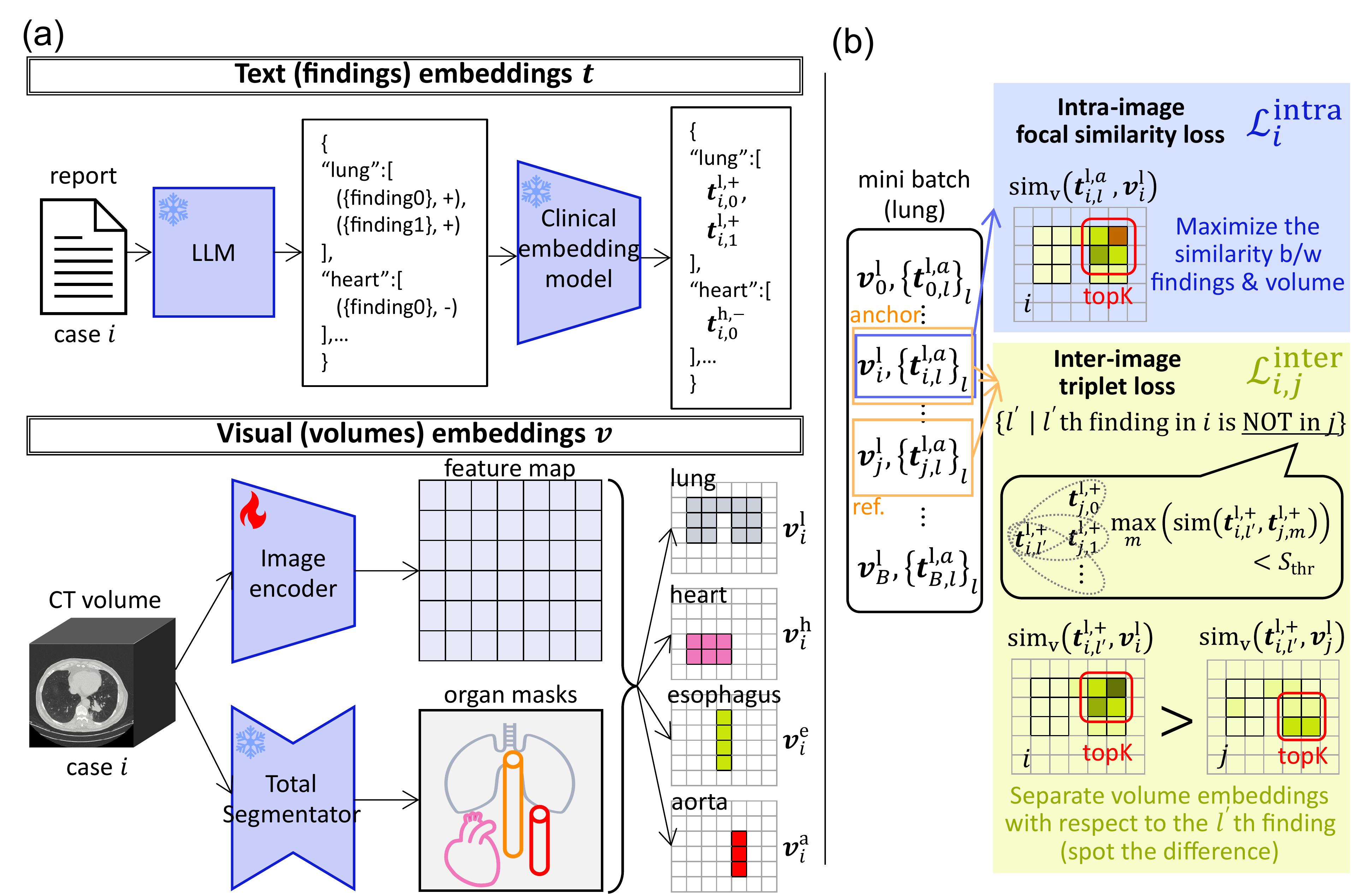}}
    \caption{Overview of the proposed strategy.
    (a) Feature extraction processes.
    See also \figref{fig:2}.
    (b) Loss function calculations.
    See Methods for details.
    }
    \label{fig:1-2}
\end{figure*}

While vision-language alignment has achieved substantial success, concerns remain when applying such approaches to medical imaging.
In CLIP-style contrastive learning, different cases are typically treated as negative pairs, meaning their feature representations are pushed apart (\figref{fig:1-1}).
Although this assumption is basically unproblematic for general-purpose image datasets spanning broad domains, domain-specific datasets often include semantically similar samples, whose representations should not be forced apart.
This issue is particularly relevant in medical imaging datasets, which often contain many clinically similar cases.
The fVLM addresses this issue by treating normal cases as equivalent and assigning them a shared feature representation.
The fVLM also estimates target similarities among abnormal cases via co-teaching, thereby drawing the representations of clinically similar cases closer together.
However, it is not straightforward to accurately estimate similarity between cases sharing similar findings because the fVLM relies solely on volume-report correspondences without additional clinical knowledge.

This study proposes a vision-language alignment learning framework that avoids treating clinically similar cases as negative pairs.
Rather than relying on a weakly supervised framework based on volume-report correspondences, we leverage the clinical knowledge embedded in pretrained language models to explicitly construct target feature representations for the image encoder.
This novel supervised framework enables the model to learn clinically appropriate representations without treating clinically similar cases as negative pairs.
Our Findings-aware Language-Image Pre-training (FiLIP) makes the following contributions:
\begin{enumerate}
    \item We propose a method to define teacher representations for the image encoder by utilizing knowledge from a large language model (LLM) and a biomedical text embedding model.
    \item We designed an intra-image focal similarity loss and a finding-conditioned inter-image contrastive loss, enabling phrase-level supervision without treating other cases as negatives solely based on case identity.
    \item FiLIP outperformed conventional weak supervision in zero-shot finding classification on both in- and external-domain evaluations, and also exhibited superior performance on downstream radiology report generation. 
\end{enumerate}

%% file: sec/2_method.tex
\section{Methods}
\subsection{FiLIP}
The core idea is to prepare teacher representations for the image encoder, realizing novel supervised learning grounded in clinical knowledge from language foundation models.
Specifically, we extracted phrases describing abnormal findings and instances, such as lesion and medical equipments, from radiology reports using an LLM, and obtained feature representations of extracted phrases using a biomedical embedding model (\figref{fig:1-1} and \figref{fig:1-2}a). 
\begin{figure*}[t]
    \centerline{\includegraphics[width=0.95\linewidth]{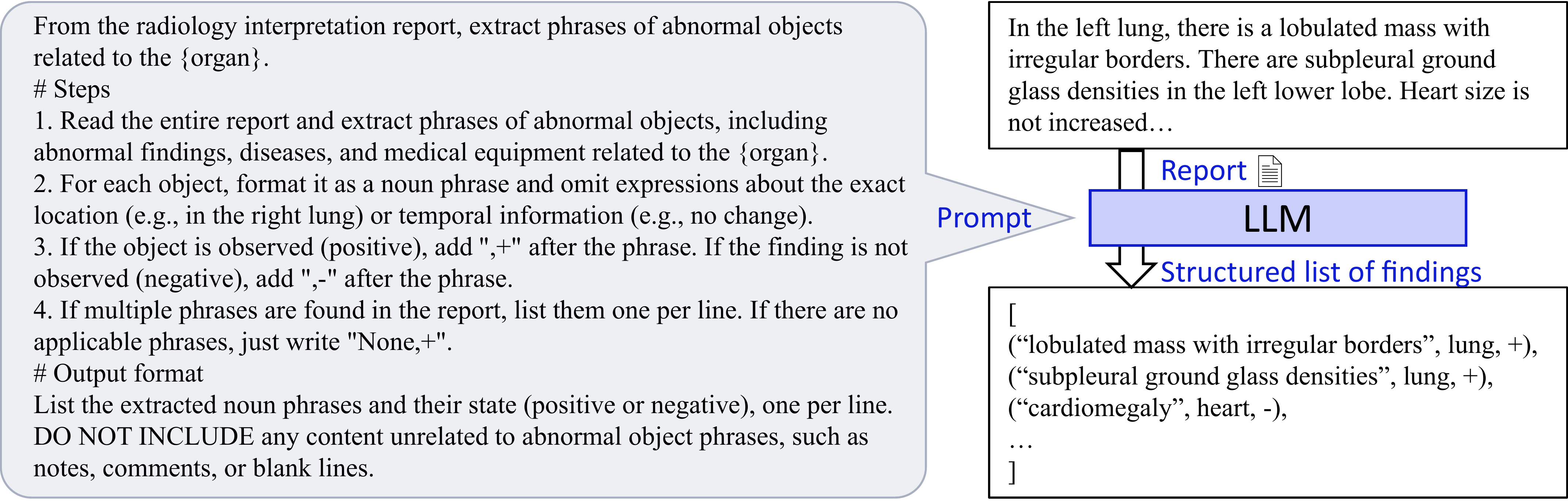}}
    \caption{Finding phrases extraction via an LLM.
    }
    \label{fig:2}
\end{figure*}

Here, the finding phrases were extracted for each predefined organ as noun phrases with modifiers, as illustrated in \figref{fig:1-1} and \figref{fig:2}.
During extraction, we removed non-visual descriptions such as temporal changes and patients' clinical background, retaining only visualized instances that the image encoder can recognize.
Besides, expressions such as ``Nodules are not observed'' were still treated as a confirmed-absent (negative) phrase ``nodules.''
Because vision-language models have difficulty learning negation \cite{Alhamoud2025}, the extracted phrases do not include explicit negation tokens such as ``no'' and ``not''.
In addition, disease-related phrases such as lung cancer may also be described with expressions of uncertainty, \eg, ``suspected'', when the radiologist identifies visual characteristics suggestive of the disease before a definitive diagnosis is established, for example by biopsy.
This reflects the fact that visual features alone are not sufficient to establish a definitive diagnosis.
In this study, phrases described with such expressions of suspicion were all treated as present (positive).
The obtained phrases thereby can include diverse modifier sequences as shown in the right bottom of \figref{fig:2}.
Each phrase was vectorized using a frozen biomedical text embedding model and used as a teacher representation for the image encoder.

The set of embedded phrases for each predefined organ $o\in \mathrm{\{lung,~heart,~esophagus,~aorta\}}$ in case $i$ is denoted by $\{\bm{t}_{i,l}^{o,a}\}_{l=0}^{N_i^{o,a}-1}$, where $a$ indicates the appearance (present: $+$ or absent: $-$), and $N_i^{o,a}$ is the number of present/absent findings, indexed by $l$.
To push apart features corresponding to background regions from those of abnormal findings, we explicitly define the background embedding shared across all cases and organs:
\begin{equation}
\label{eq:0}
\bm{t}_\mathrm{bg}=-\mathrm{Normal}\lrparen{\sum_{(i,o,a,l)\in \mathrm{Train}}\bm{t}_{i,l}^{o,a}},
\end{equation}
i.e., the vector farthest from most abnormal finding representations where $\mathrm{Normal}(\bm{v})=\bm{v}/|\bm{v}|$.
$\bm{t}_\mathrm{bg}$ was then appended to all $\{\bm{t}_{i,l}^{o,+}\}_{l=0}^{N_i^{o,+}-1}$ in the training data.

In parallel, we computed organ-wise visual features from the three-dimensional CT volume.
This process is identical to that proposed by fVLM.
We cropped the feature map obtained from the three-dimensional image encoder using the mask of organ $o$ generated by TotalSegmentator \cite{Wasserthal2023}, yielding an organ-specific volume feature map $\bm{v}_i^o$ for case $i$.

We now design novel vision-language alignment losses: an intra-image focal similarity loss and an inter-image triplet loss (\figref{fig:1-2}b).
We hypothesize that a clinically grounded image encoder can be obtained solely by training it to reproduce embeddings from a frozen language model.
The simplest way is to design a loss such that a volume's visual feature representation reproduces the list of finding embeddings $\{\bm{t}_{i,l}^{o,a}\}_{l=0}^{N_i^{o,a}-1}$ extracted from the corresponding report.
For organ $o$ in case $i$, we define the following intra-image focal similarity loss:
\begin{align}
\mathcal{L}_{i,o}^\mathrm{intra}=&-\frac{1}{N_i^{o,+}+N_i^{o,-}}\left( \sum_{l\in+}^{N_i^{o,+}}(1-p^{o,+}_{i,l})^\gamma\ln p^{o,+}_{i,l} \right. \nonumber \\
&\left. + \sum_{l\in-}^{N_i^{o,-}}(p^{o,-}_{i,l})^\gamma\ln (1-p^{o,-}_{i,l}) \right), \label{eq:1}
\end{align}
where $p^{o,a}_{i,l}$ is the prediction score based on the similarity between the text embedding and the volume feature.
A higher $p^{o,a}_{i,l}$ indicates that the volume feature likely contains the $l$-th finding, which is formulated as
\begin{align}
p_{i,l}^{o,a} &= \sigma\lrparen{ \mathrm{TopK\text{-}AvgPool}(\mathrm{sim}_\mathrm{v}(\bm{t}^{o,a}_{i,l},\bm{v}_i^o), K) } ,\label{eq:2} \\
\sigma(x) &= (1+\exp(-\alpha x+\beta))^{-1}. \label{eq:3}
\end{align}
The score $p_{i,l}^{o,a}$ reflects the relevance of the $l$-th finding to the volume, which is computed via the sigmoid function $\sigma(\cdot)$ and voxel-wise cosine similarities $\mathrm{sim}_\mathrm{v}(\cdot)$ with learnable parameters $\alpha$ and $\beta$, following ref.~\cite{Zhai2023}.
The voxel-wise cosine similarity between the finding embedding and each voxel-wise feature is aggregated via the $\mathrm{TopK\text{-}AvgPool}$ function, where the top $K\times100~\%$ of elements are selected per each organ and averaged.
This TopK average pooling is motivated because voxels relevant to a target finding are localized, and global average pooling may not be suitable.
The loss $\mathcal{L}_{i,o}^\mathrm{intra}$ encourages volume features $\bm{v}_i^o$ to be close to the present finding embedding $\bm{t}^{o,+}_{i,l}$, while pushing them away from the absent findings $\bm{t}^{o,-}_{i,l}$. 
In addition, because the target representation $\bm{t}^{o,a}_{i,l}$ may be non-uniformly distributed, we regard this as ``class imbalance'' and thus introduce the coefficients from the focal loss controlled by $\gamma$ \cite{Lin2020}.
\begin{figure}[t]
    \centerline{\includegraphics[width=0.9\linewidth]{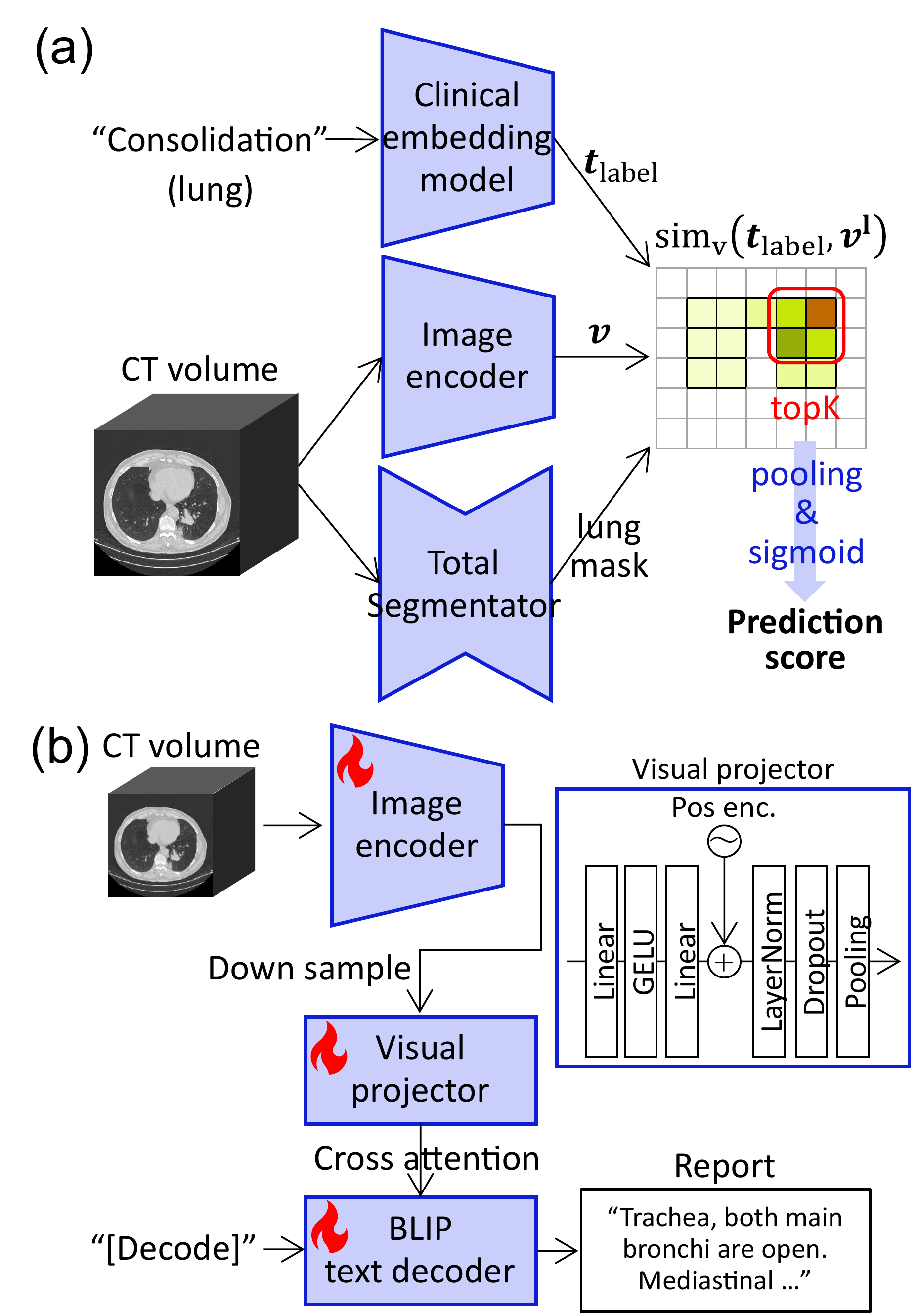}}
    \caption{Performance evaluation methods.
    (a) Zero-shot classification of finding presence.
    (b) Fine-tuning for radiology report generation as a downstream task.
    See Methods for details.
    }
    \label{fig:1-3}
\end{figure}

We additionally introduce an inter-image triplet loss that trains the image encoder in a ``spot-the-difference'' manner (\figref{fig:1-2}b).
The model trained solely with the intra-image loss may confuse visual features of coexisting findings in each single volume.
To address this concern, FiLIP provides the additional opportunities for the image encoder to distinguish between findings that frequently co-occur by comparing two volumes.
To guide spot-the-difference training, we construct a pair consisting of an anchor volume $i$ and a reference volume $j$, and define the set of unique findings present only in the anchor, $U^o_{i,j}$, as:
\begin{equation}
\label{eq:4}
U^o_{i,j} = \lrbrace{ l \left| \max_m \lrparen{\mathrm{sim}(\bm{t}^{o,+}_{i,l}, \bm{t}^{o,+}_{j,m})} < S_\mathrm{thr} \right.}, 
\end{equation}
where $\mathrm{sim}(\cdot)$ denotes scalar cosine similarity between embeddings, and $S_\mathrm{thr}$ is a hyperparameter that serves as a threshold controlling how similar two finding phrases are to be regarded as identical.
For example, in \figref{fig:1-1} bottom, ``effusion'' is included in $U$.
We then define the following binary InfoNCE-style triplet loss such that the anchor volume feature is pulled toward the embeddings of findings in $U^o_{i,j}$, while the reference volume feature is pushed away:
\begin{align}
&\mathcal{L}_{i,j,o}^\mathrm{inter} = -\frac{1}{|U^o_{i,j}|} \nonumber \\
&\sum_{l \in U^o_{i,j}}\ln\lrparen{ \frac{\exp(s(\bm{t}^{o,+}_{i,l}, \bm{v}^o_i)/\tau)}{\exp(s(\bm{t}^{o,+}_{i,l}, \bm{v}^o_i)/\tau) + \exp(s(\bm{t}^{o,+}_{i,l}, \bm{v}^o_j)/\tau)} },\label{eq:5} \\
&s(\bm{t},\bm{v}) = \mathrm{TopK\text{-}AvgPool}(\mathrm{sim}_\mathrm{v}(\bm{t},\bm{v})), \label{eq:6}
\end{align}
with a learnable temperature $\tau$.
This inter-image loss emphasizes semantic differences between volumes and, in particular, promotes the discrimination between findings that coexist within a single image.

We finally use the combination of intra- and inter-image losses as the total loss for a mini-batch $B$:
\begin{align}
\mathcal{L} = &\sum_o \left( \frac{1}{|B|}\sum_{i \in B}\mathcal{L}_{i,o}^\mathrm{intra} \right. \nonumber \\
&\left. + \frac{1}{|B|(|B|-1)}\sum_{i \in B} \sum_{j \in B, j\neq i}\mathcal{L}_{i,j,o}^\mathrm{inter} \right). \label{eq:7}
\end{align}
taking the sum over all specified organs.
Note that we train only the image encoder; the language models are kept frozen throughout this study.

As noted earlier, the objective of FiLIP is to leverage the external medical knowledge encoded in language models for training the image encoder (\figref{fig:1-1}).
Conventional weakly supervised learning based on report-volume correspondence, as in CLIP, relies on the strong assumption that different cases are clinically distinct.
However, this assumption can be substantially violated in domain-specific datasets such as medical images.
Our intra-image focal similarity loss aligns local features with predefined teacher representations aiming to train the vision-language model in a supervised rather than weakly supervised manner.
The inter-image triplet loss further separates cases only along findings on which they differ, thereby replacing case-identity-based learning with clinically grounded supervision.
The time required for inter-image triplet loss calculation and the number of phrases considered per anchor image, \ie, the effective number of negative examples per anchor, are provided in the supplementary text.

\subsection{Datasets, implementations, and experiments}
Following fVLM \cite{Shui2025} and related studies, we trained the image encoder and performed in-domain validation using an publicly available chest CT dataset, CT-RATE \cite{Hamamci2026}.
In accordance with the official data split, we used 24,114 unique cases for training and 1,557 for validation.
For external-domain testing, we used 3,630 cases from Rad-ChestCT \cite{Draelos2021}.
Preprocessing, including resolution downsampling, also followed fVLM.

We adopted a Vision Transformer (ViT) \cite{Dosovitskiy2021} as the image encoder, following fVLM.
We used GPT-4.1 \cite{Openai2025} as the LLM for structuring reports, and obtained 25,315 unique free-form finding phrases exclusively from the training reports.
We used BioLORD-2023 \cite{Remy2024} as the biomedical text embedding model, yielding 768-dims embeddings.
We expected BioLORD-2023 to provide broader specialized biomedical knowledge than both the general-purpose embedding model text-embedding-3-large \cite{Openai2024} and CXR-BERT \cite{Boecking2022}, a chest X-ray-specific model frequently used as pretrained weights for medical vision-language models; this expectation was subsequently verified through ablation studies.
For training, we adopted the hyperparameters from fVLM official implementation \cite{Zhang2026}, such as the learning rate and batch size.
The FiLIP-specific learnable parameters $\alpha$, $\beta$, and $\tau$ were initialized to 4, 2, and 0.07, respectively.
The hyperparameters $K$, $\gamma$, and $S_\mathrm{thr}$ were set to 0.75, 2, and 0.6, respectively, and we evaluated performance dependence on them via ablation studies.

\begin{figure*}[t]
  \centering
  \includegraphics[width=\linewidth]{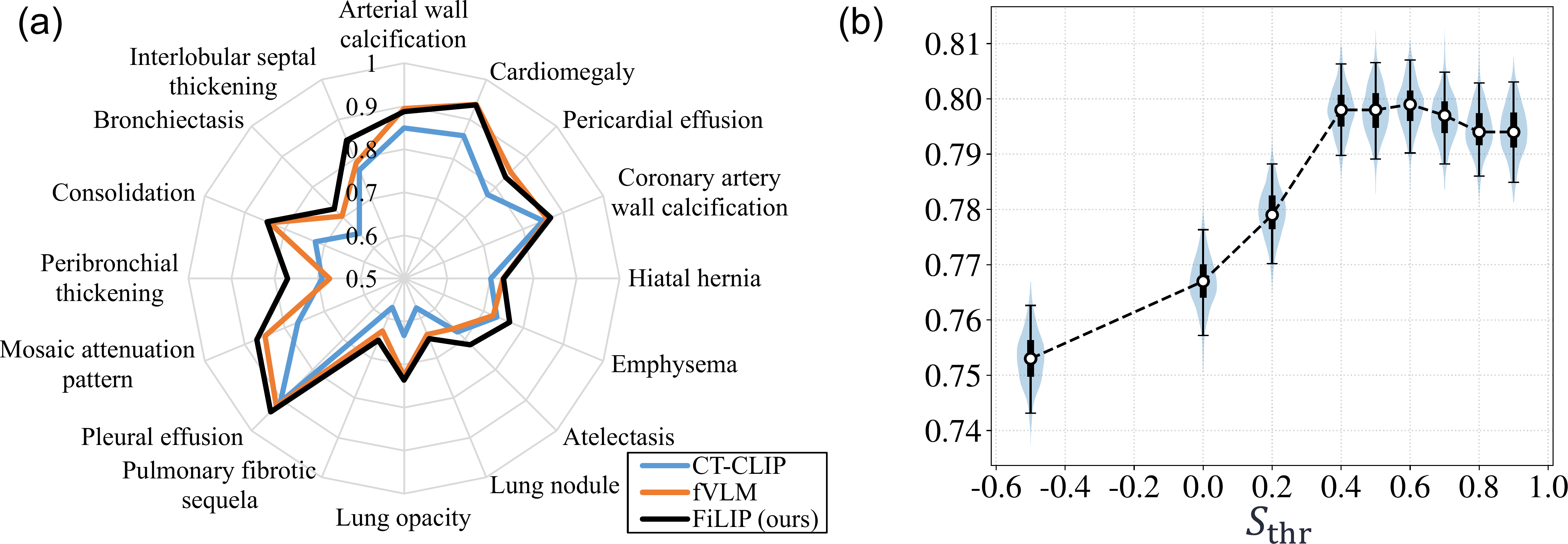}
  \caption{Zero-shot finding classification performance for CT-RATE.
  (a) AUC for each finding. See the supplementary text for correspondence between labels and organs.
  (b) Dependency of the mean AUC on the similarity threshold $S_\mathrm{thr}$ for computing $\mathcal{L}^\mathrm{inter}$.
  Violin plots show the AUC distributions, with bootstrap-estimated 95\% confidence intervals and interquartile ranges overlaid.}\label{fig:3}
\end{figure*}

We evaluated the performance of the proposed method by a zero-shot classification of finding presence in each CT volume.
Specifically, for each of the 16 findings defined in the fVLM evaluation protocol \cite{Shui2025}, we calculated the area under the ROC curve (AUC).
During inference, we computed embeddings of each label term using BioLORD-2023 and performed classification by taking the similarity with the volume feature map of the organ where the finding occurs; the score $p$ in \equaref{eq:2} was used as the probability of the finding being present (see \figref{fig:1-3}a for the case of the consolidation in lungs).
To demonstrate the advantage of FiLIP over existing loss functions for vision-language alignment, we compared the zero-shot finding classification performance against recently proposed CT-CLIP \cite{Hamamci2026}, fVLM \cite{Shui2025}, and SimCroP \cite{Wang2025}, which have similar model architectures and are therefore suitable for comparison.
The baseline methods were re-evaluated using their published weights trained on CT-RATE, with confidence intervals (CI) computed for fair comparison.
For SimCroP, note that because a zero-shot inference method is not officially provided, the cosine similarity between the globally averaged volume feature and the text embedding of the prompt ``\{finding\} is present.'' was leveraged as the score, following the training strategy reported in ref.~\cite{Wang2025}.
The classification performance was evaluated by averaging the AUC across all finding classes.
The 95\% CI of the mean AUC were determined via a non-parametric bootstrap procedure with 1,000 resamples, and statistical significance was assessed at a 5\% significance level using two-sided DeLong tests.
Besides, we also compared the AUCs by CT-CLIP variants fine-tuned with class labels (VocabFine and ClassFine) and Merlin \cite{Blankemeier2026} evaluated in the same setup and reported in ref.\cite{Cao2025}.
Note that these models are not directly comparable because they were trained with additional data beyond radiology reports.
\begin{table}[t]
  \centering
  \caption{The mean AUC for zero-shot finding classification. 95\% confidence intervals are shown in the brackets. $\dagger$ indicates models requiring additional data beyond radiology reports for training, and values are from ref.\cite{Cao2025}.
  FiLIP significantly outperformed fVLM on both evaluation datasets (see the main and supplementary text for details).}\label{tb:1}
  \small
  \begin{tabular}{l||l@{\hspace{6pt}}l}
      & \multicolumn{2}{c}{Evaluation dataset} \\
    Model & CT-RATE & Rad-ChestCT\\
    \hline\hline
    CT-CLIP \cite{Hamamci2026} & 0.733 [0.727-0.740] & 0.632 [0.625-0.640] \\
    \hspace{1em} (VocabFine)$^\dagger$ & 0.760 & 0.657 \\
    \hspace{1em} (ClassFine)$^\dagger$ & 0.761 & 0.647 \\
    Merlin$^\dagger$ \cite{Blankemeier2026} & 0.728 & 0.644 \\
    fVLM \cite{Shui2025} & 0.778 [0.770-0.786] & 0.680 [0.674-0.686] \\
    SimCroP \cite{Wang2025} & 0.620 [0.613-0.633] & 0.509 [0.505-0.521] \\
    {\bf FiLIP (ours)} & {\bf 0.799 [0.791-0.808]} & {\bf 0.690 [0.683-0.697]}
  \end{tabular}
\end{table}
\begin{table}[t]
  \centering
  \caption{Ablation studies on losses evaluated using CT-RATE.}\label{tb:2}
  \small
  \begin{tabular}{ccccc|c}
      $\mathcal{L^\mathrm{intra}}$ & $\bm{t}_\mathrm{bg}$ & TopK & Focal & $\mathcal{L^\mathrm{inter}}$ & AUC \\
      \hline\hline
      $\checkmark$ & & & & & 0.761 \\
      $\checkmark$ & $\checkmark$ & & & & 0.764 \\
      $\checkmark$ & $\checkmark$ & $\checkmark$ & & & 0.768 \\
      $\checkmark$ & $\checkmark$ & & $\checkmark$ & & 0.774 \\
      $\checkmark$ & $\checkmark$ & $\checkmark$ & $\checkmark$ & & 0.779 \\
      & & & & $\checkmark$ & 0.788 \\
      $\checkmark$ & $\checkmark$ & & & $\checkmark$ & 0.797 \\
      $\checkmark$ & $\checkmark$ & $\checkmark$ & $\checkmark$ & $\checkmark$ & {\bf 0.799}
  \end{tabular}
\end{table}
\begin{table}[t]
  \centering
  \caption{Ablation studies on foundation models evaluated using CT-RATE.
  \# 4 is the default setting in this study.
  Statistical significance was observed between \# 1 vs \# 4, and \# 2 vs \# 4, while no significant difference was found between \# 3 vs \# 4 (see the main and supplementary text for details).
  }\label{tb:2.1}
  \footnotesize
  \begin{tabular}{lll|c}
       & LLM & embedding model & AUC \\
      \hline\hline
      \# 1 & GPT-4.1 & CXR-BERT & 0.744 [0.734-0.754] \\
      \# 2 & GPT-4.1 & text-embedding-3-large & 0.787 [0.778-0.795] \\
      \# 3 & gpt-oss-120B & BioLORD-2023 & {\bf 0.796 [0.787-0.804]} \\
      \# 4 & GPT-4.1 & BioLORD-2023 & {\bf 0.799 [0.791-0.808]}
  \end{tabular}
\end{table}

To verify that the FiLIP-trained model comprehensively captures clinical features beyond finding presence, we also evaluated its performance on a downstream task.
Specifically, we trained an end-to-end model for radiology report generation using CT-RATE.
The pretrained image encoder was connected to the BLIP text decoder \cite{Li2022} through cross-attention mediated by a visual projector, and the entire network was fine-tuned (\figref{fig:1-3}b, see the supplementary text for details).
Performance was evaluated using lexical metrics (BLEU \cite{Papineni2002}, METEOR \cite{Banerjee2005}, and ROUGE \cite{Lin2004}) and the GREEN score \cite{Ostmeier2024}, a LLM-based metric that accounts for clinical semantic similarity between texts.
In addition to overall score, the GREEN analyzes the following error/match counts: false reporting of a finding in the candidate report, omission of a finding present in the reference report, misidentification of a finding’s anatomic location or position, misassessment of a finding’s severity, inclusion of a comparison not described in the reference report, omission of a comparison describing interval change from a prior study, and matched findings.
Because the report generation model in this study analyzes a single CT volume, evaluation categories related to longitudinal comparison were not considered.
For each metric, uncertainty was estimated using the standard error.
For the BLEU-4 and the GREEN including subcategories, statistical significance of differences between FiLIP and prior methods was assessed at a 5\% significance level using two-sided Wilcoxon signed-rank tests.
All p-values in this study were adjusted for multiple comparisons according to the Holm correction.
\begin{table*}[t]
  \centering
  \caption{Results on the downstream task of report generation using CT-RATE.
  Standard errors are shown in the brackets.
  For FiLIP scores, *1, *2, and *3 indicate statistically significant differences from ImageNet, fVLM, and SimCroP, respectively.
  *0 indicates no significant difference from any model. Metrics without an significance marker were not statistically tested. See p-values for the supplementary text.}
  \label{tb:3}
    \begin{minipage}{\linewidth}
    \centering
    \captionof*{table}{(a) Evaluation using lexical metrics.}
    {
    \small
    \begin{tabular}{l||cccccccc}
      Model & BLEU-1 & BLEU-2 & BLEU-3 & BLEU-4 & METEOR & ROUGE-1 & ROUGE-2 & ROUGE-L\\
      \hline\hline
      ImageNet & 0.3794(38) & 0.2852(38) & 0.2258(37) & 0.1771(34) & 0.3903(34) & 0.5816(38) & 0.3832(57) & 0.4483(53) \\
      fVLM \cite{Shui2025} & 0.3765(38) & 0.2824(39) & 0.2239(37) & 0.1759(35) & 0.3889(35) & 0.5797(40) & 0.3823(58) & {\bf 0.4485(54)} \\
      SimCroP \cite{Wang2025} & 0.3915(38) & 0.2925(39) & {\bf 0.2307(39)} & {\bf 0.1810(37)} & {\bf 0.3975(37)} & 0.5886(39) & 0.3815(58) & 0.4444(54) \\
      {\bf FiLIP (ours)} & {\bf 0.3938(34)} & {\bf 0.2934(36)} & 0.2306(36) & 0.1798(33)${}^{*1,2}$ & 0.3974(33) & {\bf 0.5893(37)} & {\bf 0.3837(56)} & 0.4472(52)
    \end{tabular}
    }
    \end{minipage}
    
    \vspace{1em}
    
    \begin{minipage}{\linewidth}
    \centering
    \captionof*{table}{(b) Evaluation using the clinical metric, the GREEN score. The subcategories represent the counts of following subjects: g-1, false report of a finding in the candidate; g-2, missing a finding present in the reference; g-3, misidentification of a finding's anatomic location/position; g-4, misassessment of the severity of a finding; g-5, matched findings.
    }
    {
    \small
    \begin{tabular}{l||c|ccccc}
        &  & \multicolumn{5}{c}{subcategory} \\
      Model & overall & g-1 ($\downarrow$) & g-2 ($\downarrow$) & g-3 ($\downarrow$) & g-4 ($\downarrow$) & g-5 ($\uparrow$) \\
      \hline\hline
      ImageNet & 0.5416(67) & 0.982(31) & {\bf 3.376(60)} & 0.1019(85) & 0.0679(68) & 6.13(11) \\
      fVLM \cite{Shui2025} & 0.5426(68) & 0.969(29) & 3.381(60) & 0.0936(81) & 0.0724(69) & 6.23(11) \\
      SimCroP \cite{Wang2025} & 0.5464(67) & 0.883(31) & 3.383(58) & 0.1091(97) & 0.0842(75) & 6.19(11) \\
      {\bf FiLIP (ours)} & {\bf 0.5533(68)}${}^{*1,2}$ & {\bf 0.747(27)}${}^{*1,2,3}$ & 3.387(57)${}^{*0}$ & {\bf 0.0641(65)}${}^{*1,2,3}$ & {\bf 0.0615(64)}${}^{*0}$ & {\bf 6.30(11)}${}^{*0}$
    \end{tabular}
    }
    \end{minipage}

\end{table*}


%% file: sec/3_results.tex
\section{Results}
The mean AUC of zero-shot finding classification is reported in \tabref{tb:1}.
Our model outperformed existing models trained with conventional CLIP-style methods on both the in-domain validation and the external-domain test for zero-shot finding classification.
While CIs are overlapped for the external test, the improvement was shown to be statistically significant for both datasets by DeLong tests (see the supplementary text for p-values).
A per-finding comparison is also shown in \figref{fig:3}a.
Classification performance improved particularly for lung findings, and we did not observe substantial degradation in AUC for any specific finding.
As noted in the original paper \cite{Wang2025}, SimCroP does not execute instance-level alignment for the entire volume and therefore showed relatively lower classification performance, while it demonstrated strong performance on a downstream task as described later.

The validity of the proposed method was further underpinned by ablation studies (\tabref{tb:2} and \figref{fig:3}b).
We observed performance drop when removing $\bm{t}_\mathrm{bg}$, replacing the focal loss with a cross-entropy loss, or replacing TopK average pooling with global average pooling in $\mathcal{L^\mathrm{intra}}$.
Remarkably, using $\mathcal{L^\mathrm{intra}}$ alone, i.e., using only correspondences within the same case, still achieved performance comparable to fVLM.
We also found that training with only the inter-image triplet loss $\mathcal{L^\mathrm{inter}}$ outperformed fVLM.
The best performance was realized by combining $\mathcal{L^\mathrm{intra}}$ and $\mathcal{L^\mathrm{inter}}$.
When varying the similarity threshold $S_\mathrm{thr}$ to determine the set of unique cases $U$, we observed peak performance around $S_\mathrm{thr} \approx 0.6$.
This is consistent with the following intuition: If $S_\mathrm{thr}$ is set too high or too low, common findings between anchor and reference volumes cannot be identified reliably, which in turn hampers the spot-the-difference training.
The plateau near the peak further suggests that our method is not highly sensitive to the choice of the hyperparameter $S_\mathrm{thr}$.
Moreover, we observed that the performance was significantly affected by the choice of the text embedding model (\tabref{tb:2.1}), indicating that FiLIP successfully utilized the clinical knowledge embedded in BioLORD.
We also confirm that the performance was not substantially decayed when replacing GPT-4.1 with gpt-oss-120B \cite{Openai2025oss}, showing that models can be trained in private environments as well (see p-values for the supplementary text).

\tabref{tb:3} summarizes the fine-tuning results for radiology report generation with pretrained image encoders.
We compared a ViT with attention weights pretrained on ImageNet with those trained with fVLM, SimCroP, and FiLIP.
While FiLIP exhibited no significant difference to SimCroP regarding conventional lexical metrics, significant difference was observed regarding the GREEN, which measures the clinical semantic similarity beyond word-level overlap and has been reported to correlate well with radiologists' assessments \cite{Ostmeier2024, Sharma2026} (see p-values for the supplementary text).
Across the subcategories, FiLIP significantly reduced false reporting of findings (g-1) but also improved the localization of findings (g-3) compared to SimCroP.
The overall GREEN score and BLEU-4 was significantly higher for FiLIP compared to fVLM.
These results suggest that the FiLIP-trained encoder learned to capture not only features required for classification of finding presence but also diverse clinical information to reproduce radiology reports.
\begin{figure*}[t]
  \centering
  \includegraphics[width=0.9\linewidth]{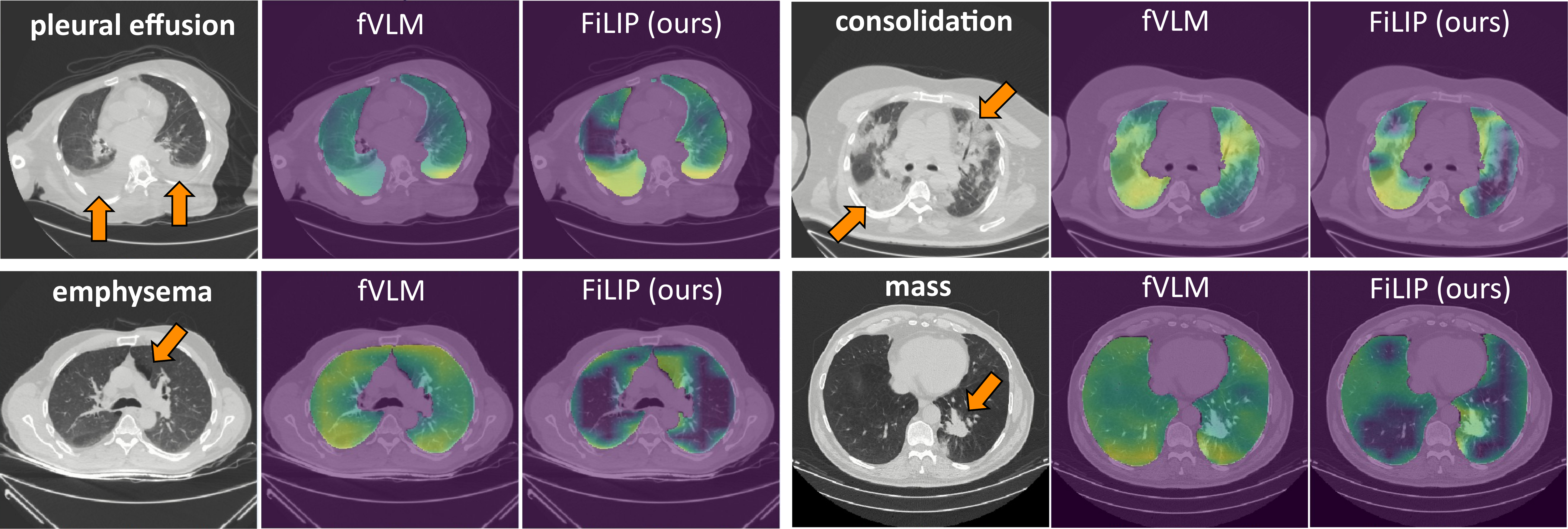}
  \caption{Visualization of voxel-wise similarity map between target finding embeddings and volume features.
  The arrows indicate the position of each finding.
  The brighter color corresponds to the higher similarity.}\label{fig:4}
\end{figure*}

To visually assess the improvement in the g-3 score, we observed the zero-shot capability of finding localization by the FiLIP-trained model.
For models trained with fVLM and FiLIP, we visualized pseudo-color maps of the cosine similarity between voxel-wise features and the each finding embedding, normalized using the minimum and maximum values over the entire volume (\figref{fig:4}).
Consolidation and pleural effusion were recognized relatively well by both models (\figref{fig:3}a).
In contrast, for emphysema, the fVLM-trained model produced an almost uniform similarity map, whereas FiLIP yielded an anatomically plausible map.
Our model also exhibited a spatially coherent map correlated with mass, which was not included in the zero-shot finding classification evaluation.
Note that quantitative evaluation is required in future studies to prove superior performance in zero-shot finding localization.

%% file: sec/4_conclusions.tex
\section{Discussions and Conclusions}
From a clinical perspective, the following limitations remain.
Although we demonstrated improved zero-shot finding classification, more comprehensive zero-shot prediction remains to be investigated, regarding information beyond findings that can be omitted during LLM-based extraction.
Even for finding classification, the performance gain for lung nodules, one of the most critical findings relevant to cancer, is relatively modest.
Consistently, the similarity map for the cancer-related mass (\figref{fig:4}) exhibits a false positive near the slightly brighter region in the right lung (left in the image).
To address this issue, since cancer-related findings are often highly localized, the optimal $K$ may also be smaller for such findings.
It may be possible to adaptively select visual tokens for each finding, as in ref.~\cite{Gorti2022}.
Regarding the downstream report generation, no significant improvement from SimCroP was observed in the overall GREEN and several subcategories.
Severity assessment may require more expertised clinical knowledge.
In addition, the reference reports used for evaluation include information missed in the chest CT volume, such as head, neck, and abdomen findings and patient background, which may make score differences less detectable.
Future work should consider preprocessing the reference reports and obtaining more precise evaluation by radiologists.

In conclusion, we propose supervised vision-language alignment, motivating a rethinking of the common strategy for multimodal medical representation learning.
By leveraging the knowledge encoded in the LLM and the biomedical text embedding model to define teacher representations for the image encoder, our approach outperforms prior vision-language alignment methods for zero-shot finding classification and downstream radiology report generation.
Experiments on downstream report generation suggest that the FiLIP-trained encoder captures features related not only to the presence of findings but also to their location.
In other words, FiLIP has the potential to learn comprehensive clinical features contained in CT volumes.
Regarding the hyperparameters, since the value $\gamma=2$ employed in the experiments is the standard choice for focal loss, only two hyperparameters, $K$ and $S_\mathrm{thr}$, need to be tuned in practice.
The ablation study showed that the performance variations associated with $K$, \ie, the TopK pooling parameter, and $S_\mathrm{thr}$ were limited, suggesting that the burden of hyperparameter search is not substantial.

The performance of the FiLIP-trained model may further improve as language models continue to advance.
As the biomedical embedding models improve, image encoders may learn to capture increasingly fine-grained clinical feature representations.
For report structuring, improvements in phrase extraction are also expected as LLMs evolve.
Further data augmentation is also possible, \eg, by using external medical knowledge to supplement missing relationships among phrases. 
In addition, FiLIP can be combined with any network architecture and auxiliary losses, such as image reconstruction and report generation objectives, as explored for ViSD-Boost \cite{Cao2025}, SimCroP \cite{Wang2025}, and CORIPLI \cite{Wald2025}.
Questions remain regarding how to construct the list of teacher representations, including how to define $\bm{t}_\mathrm{bg}$ (\equaref{eq:0}) and incorporate negative phrases not mentioned in the report into training.
FiLIP has the potential for scalability and transferability to other organs and modalities, as performance decay is modest across hyperparameter choices, possibly encouraging broader investigation and adoption in domain-specific vision-language alignment beyond the medical field.